\pdfoutput=1

\documentclass[11pt]{article}

\usepackage[final]{acl}

\usepackage{times}
\usepackage{latexsym}
\usepackage{booktabs}
\usepackage{xspace}

\usepackage{xcolor}
\def\eg{\emph{e.g.}} 
\def\ie{\emph{i.e.}} 
 
\def\etc{\emph{etc.}} \def\vs{\emph{vs.}\xspace}

\definecolor{Highlight}{HTML}{2a9d8f}
\definecolor{Highlight2}{HTML}{cc3300}
\definecolor{LightHighLight}{HTML}{cbf3f0}
\definecolor{mygray}{HTML}{e9ecef}

\usepackage{multirow}
\usepackage{array}
\usepackage{makecell}
\usepackage{amsmath}
\usepackage{amsfonts}

\newcommand{\caml}{CaMML\xspace}

\usepackage{colortbl}

\usepackage{graphicx}

\usepackage[T1]{fontenc}

\usepackage[utf8]{inputenc}

\usepackage{microtype}

\usepackage{inconsolata}

\title{CaMML: Context-Aware Multimodal Learner for Large Models}

\author{Yixin Chen$^*$$^\dagger$ \\
  The Chinese University of Hong Kong \\
  \texttt{yxchen@cse.cuhk.edu.hk} \\\And
  Shuai Zhang$^*$ \\
  Amazon Web Services \\
  \texttt{shuaizs@amazon.com} \\\AND
  Boran Han \\
  Amazon Web Services \\
  \texttt{boranhan@amazon.com} \\\And
  Tong He \\
  Amazon Web Services \\
  \texttt{htong@amazon.com} \\\And
  Bo Li$^\dagger$ \\
  The University of Chicago \\
  \texttt{bol@uchicago.edu}
  }

\begin{document}
\maketitle

\def\thefootnote{$^*$}\footnotetext{Co-first Author}\def\thefootnote{\arabic{footnote}}
\def\thefootnote{$^\dagger$}\footnotetext{Work done at Amazon}\def\thefootnote{\arabic{footnote}}

\begin{abstract}

In this work, we introduce \textbf{C}ontext-\textbf{A}ware \textbf{M}ulti\textbf{M}odal \textbf{L}earner (\caml), for tuning large multimodal models (LMMs). \caml, a lightweight module, is crafted to seamlessly integrate multimodal contextual samples into large models, thereby empowering the model to derive knowledge from analogous, domain-specific, up-to-date information and make grounded inferences. Importantly, \caml is highly scalable and can efficiently handle lengthy multimodal context examples owing to its hierarchical design. Based on \caml, we have developed two multimodal models, \caml-7B and \caml-13B, that have shown exceptional performance across an array of benchmark datasets for multimodal tasks. Remarkably, \caml-13B achieves the state-of-the-art performance on over ten widely recognized multimodal benchmark datasets, surpassing LLaVA-1.5 (13B) with a noticeable margin, without integration of any external resources. Moreover, we have conducted extensive ablative studies to inspect the inner workings of \caml and performed qualitative analyses to showcase its effectiveness in handling real-world challenging cases. Code and models are available at: \url{https://github.com/amazon-science/camml}.

\end{abstract}

\section{Introduction}

\begin{figure}[t]
	\centering
		\includegraphics[width=1.0\linewidth]{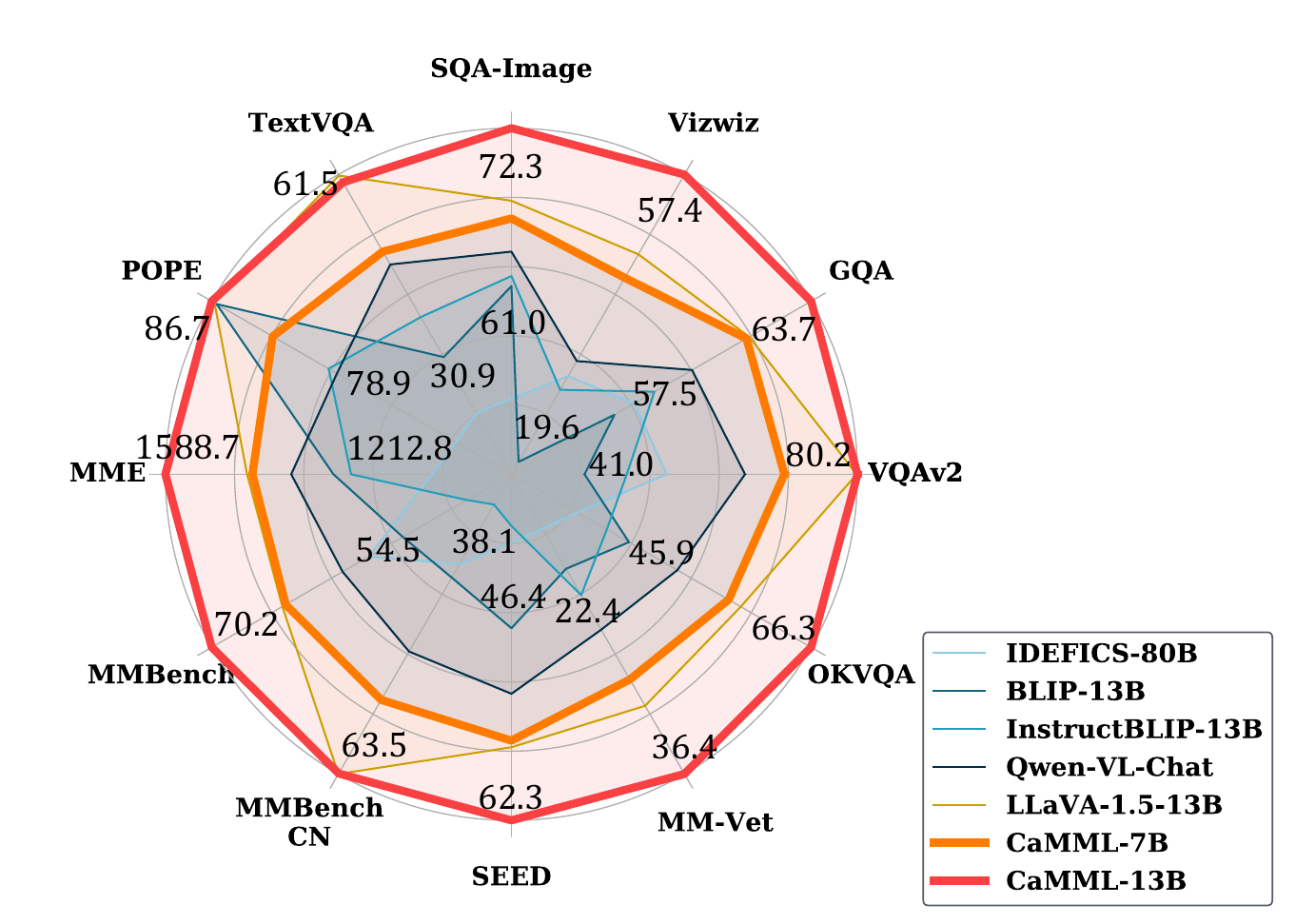}
	\vspace{-0.2in}
	\caption{\caml achieves the state-of-the-art performance on a number of multimodal benchmarks, outperforming LLaVA-1.5 and many other large multimodal models.}
    \vspace{-0.2in}
    \label{fig:caml_teaser_fig}
\end{figure}

Recently, large multimodal models (LMMs)~\cite{llava,llamaadapter,ofa,llava15, minigpt4, frozen, flamingo} have demonstrated remarkable performance in a variety of tasks, including but not limited to  visual question answering, image captioning, visual grounding, visual-language reasoning, optical character recognition, and visual entailment. Notably, in certain benchmark assessments, these multimodal foundation models have even exceeded human-level performance~\cite{mmcot,llava}. 

Despite the impressive performance, their ability to make inferences is constrained by the knowledge encoded in the model parameters. The inflexible design of these models makes it challenging for them to generalize from contextual examples. For instance, LLaVA-1.5 falls short in processing multiple images and attributes it to the lack of corresponding instruction-tuning training data~\cite{llava15}. Nevertheless, learning and making inferences through contextual examples are fundamental elements of our cognitive processes. Human beings frequently tackle intricate problems by relying on past experiences and identifying analogous situations. Taking inspiration from the cognitive process, we hypothesize that empowering large multimodal models with the capability to perceive and derive insights from analogous contextual examples can significantly streamline the inference process and lead to more precise predictions. Nonetheless, the means to replicate the human cognitive processes in LMMs remain unclear.

As such, our goal is to empower multimodal foundational models to harness context-aware learning, thereby enhancing their ability to comprehend and adapt to previously unseen examples. Identifying relevant context examples is relatively straightforward; this process can be facilitated using multimodal embedding models like ImageBind~\cite{imagebind} or CLIP~\cite{clip}. This approach simulates the act of recalling similar situations from past experiences. Yet, effectively and efficiently integrating the identified similar context samples into large models poses challenges, particularly given the potential variability in the number of context samples and interleaved modalities, resulting in lengthy and heterogeneous context input.

To this end, we propose a context-aware multimodal learner, dubbed as \caml, for LMMs. 
\caml acts as a crucial intermediary between the contextual examples and a large language model (LLM). Our approach is structured hierarchically, where the initial level establishes connections between the text and image modalities for each example through cross-attention mechanisms. This integration of text and image information enables a deeper understanding of the interleaved context. Following this, another module takes the outputs of the first level and performs cross-attention between the contextual information and a predefined set of learnable, fixed-length tokens. The resulting output from this level is then used as fixed-length input for the LLM, allowing the model to leverage the refined and context-aware information to perform complex multimodal understanding tasks. To summarize, we make the following contributions: 
\begin{itemize}
    \item  We propose \caml, a context-aware multimodal learning approach for finetuning multimodal models. \caml is lightweight and can be applied to process extremely long multimodal context samples; 
    \vspace{-0.1in}
    \item With \caml, we have developed two multimodal models, \caml-7B and \caml-13B. These models have achieved state-of-the-art performance across a diverse range of benchmarks encompassing various multimodal tasks, all without the need for external data integration; 
    \vspace{-0.1in}
    \item We conduct comprehensive model analyses and case studies to examine the internal mechanisms of \caml and showcase how the proposed model can effectively handle real-world challenging cases. 
\end{itemize}

\section{Related Work}

\paragraph{Large Multimodal Models} The success of LLMs has sparked a burgeoning interest in scaling up multimodal models. One prevalent strategy involves the incorporation of vision encoders, such as ViT~\cite{vit} and CLIP, into existing LLMs (\eg, LLaMA~\cite{llama}, Vicuna~\cite{vicuna}). For example, LLaMA-adapter~\cite{llamaadapter} inserts learnable adaption prompts into LLaMA and make it to follow multimodal instructions for multimodal reasoning. BLIP2~\cite{blip2} bridges different modalities via Q-Former and employs a two-stage training method to bootstrap both representation learning and LLM (\eg, OPT~\cite{opt}, Flan-T5~\cite{flant5}) generative learning capabilties. LLaVA~\cite{llava, llava15} projects encoded image features to text token space using linear layers and shows state-of-the-art performance on a variety of multimodal tasks. Another noteworthy series of strategy is to unify multimodal input data and pretrain the model from scratch, exemplified by OFA~\cite{ofa}, Perceiver~\cite{perceiver, perceiverio}, Uni-Perceiver~\cite{uniperceiver}, Unival~\cite{unival}, and Unified-IO~\cite{unifiedio}. These models map images, text, and other modalities into the same IO space and use a causal language model objective for training. 

\vspace{-0.1in}
\paragraph{Multimodal Few-shot Learning} Extensive research has been conducted on learning from a limited number of multimodal examples. Among them, Flamingo~\cite{flamingo} introduces gated xatten-dense layers to establish cross-modal interactions between visual input and text input for few-shot learning. Frozen~\cite{frozen} trains a vision encoder to produce a sequence of image embeddings and input it to frozen language models for multimodal tasks. An alternative approach for tackling multimodal few-shot learning involves leveraging retrieval augmented generations (RAG)~\cite{lewis2020retrieval,borgeaud2022improving,ram2023context,khandelwal2019generalization}. This technique enables the models to access external knowledge repositories, databases, or structured data sources, allowing them to access the up-to-date as well as domain-specific expertise when crafting responses. Recently, there has been a surging interest for harnessing RAG within the realm of multimodal models~\cite{murag,revilm,racm3,liu2023learning}. Amalgamating RAG with multimodal models helps provide contextually-rich responses and opens up exciting possibilities across domains such as image captioning~\cite{revilm}, image generation~\cite{racm3}, etc. Among them, \citet{murag} propose MuRAG, a model of size 527M, for multimodal-QA. MuRAG concatenates the visual embeddings (extracted by ViT) and text word embeddings of retrieved multimodal image-text pairs and requires the language model to handle a very lengthy input sequences.  Re-ViLM~\cite{revilm} is tailored for image captioning and it follows the Flamingo architecture and only uses the retrieved text for augmentation. RA-CM3~\cite{racm3} leverages retrieval to augment the CM3~\cite{cm3} pipeline which can do both text-to-image and image-to-text generations.

In contrast to previous approaches, our primary objective is to develop a lightweight module capable of processing multimodal context information efficiently and effectively, with superior generalization capabilities. This novel approach enables LLM to proficiently reason from multimodal in-context examples, leading to more accurate and precise inferences. As discussed in \cite{llava15}, LLaVA-1.5 faces limitations when confronted with scenarios involving multiple images and lengthy contexts. \caml can effectively address these challenges, a fact further corroborated by its superior performances across multiple benchmarks.

\section{Context-Aware Multimodal Learner}

\begin{figure*}[t]
    \centering
    \includegraphics[width=1.0\linewidth]{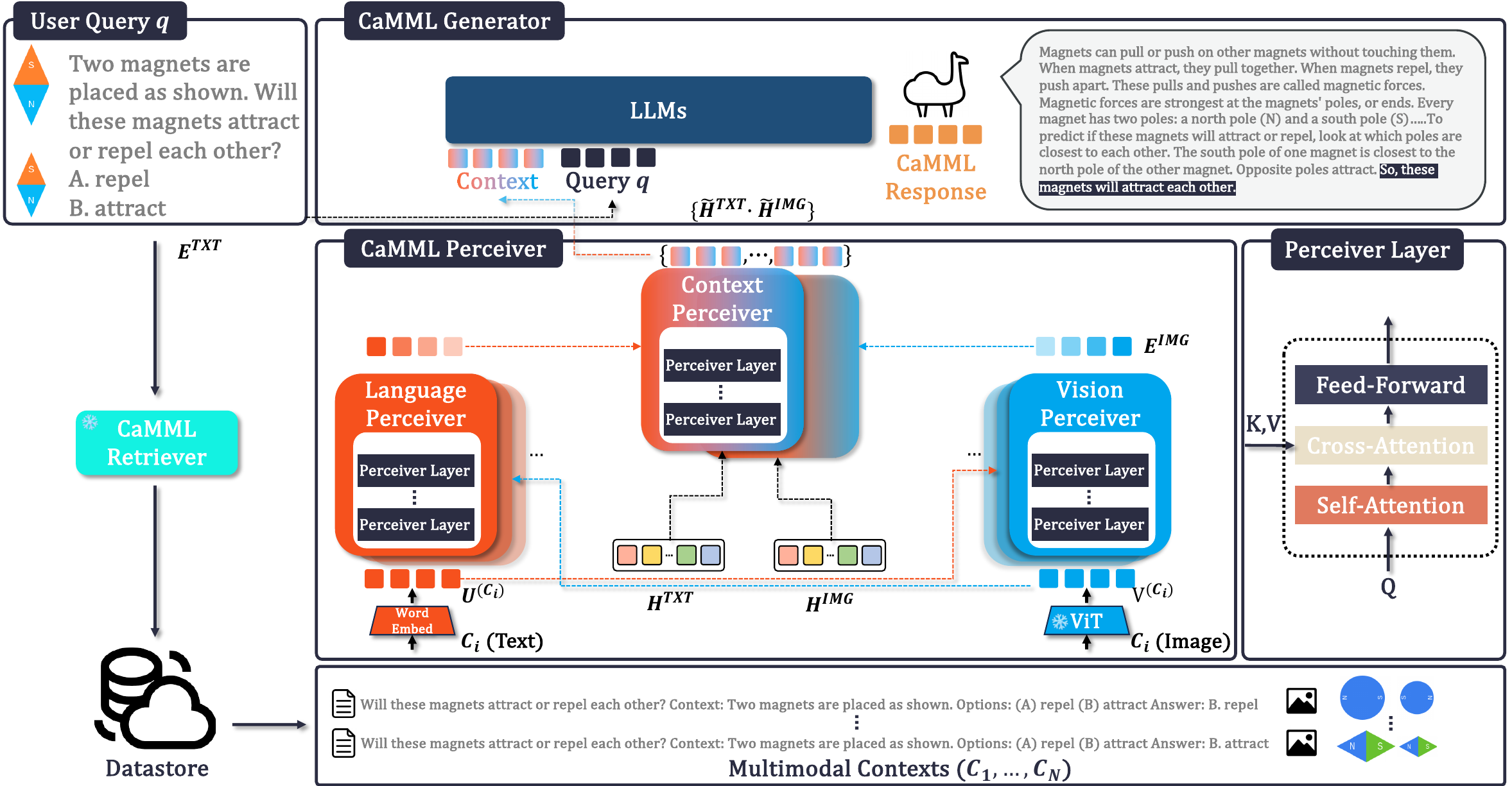}
    \vspace{-0.2in}
    \caption{\caml framework, which consists of retriever, perceiver and generator. Once receiving user query $q$, \caml retriever identifies relevant multimodal contexts $C$ from datastore, then \caml Perceiver seamlessly integrates various modalities, effectively encodeing long-context information and injecting it into the \caml generator. This allows for the prediction of responses that are conditioned on both the context and the query $q$.}
    \vspace{-0.15in}
    \label{fig:arch}
\end{figure*}

\subsection{Architecture of \caml}
The structure of \caml is depicted in Figure \ref{fig:arch}. We provide a detailed explanation of each component below.

\subsubsection{Datastore and Context Retriever} The datastore is created from either the training set or external resources. We adopt an embedding encoder, ImageBind~\cite{imagebind}, to extract dense vector representations for every multimodal sample contained in the datastore. Following this, we build an index using Faiss~\cite{johnson2019billion}, a highly efficient similarity search library, to enable rapid and efficient search operations. During both the training and inference phases, our approach involves identifying the top $N$ most closely related multimodal samples, denoted as $C_1, C_2, \ldots, C_N$, from the datastore using Faiss, based on a given multimodal query $q$\footnote{Specifically, we compute the similarity between query text and context image, and top-k images along with their corresponding texts are recalled from the datastore.}. It is worth noting that the retriever remains frozen throughout all stages of the process.

\subsubsection{Multimodal \caml Perceiver} To seamlessly integrate interleaved multimodal samples, we introduce a novel module denoted as \caml Perceiver, which revolves around two key design principles: (1) The module should accommodate a dynamic number of context samples without imposing a significant computational burden. It is worth noting that directly concatenating the image tokens and text tokens will result in $L = \sum^N_{i=1} (T_{i}^{\text{IMG}} + T_{i}^{\text{TXT}})$ tokens, where $T_{i}^{\text{IMG}}$ is the number of image tokens and $T_{i}^{\text{TXT}}$ is the number of text tokens. The formulation results in a linear increase of $L$ with respect to $N$, posing scalability concerns; (2) It is crucial that the text and image in each individual sample remain tightly coupled. 

To achieve this, we embrace a hierarchical design, facilitating the transformation of the $N$ context samples into $M$ tokens, where $M < L$. Formally, the multimodal \caml Perceiver comprises three key modules: a \textit{Vision Perceiver} (VP) denoted as $f_{\theta^{\text{VP}}}$ , a \textit{Language Perceiver} (LP) denoted as $f_{\theta^{\text{LP}}}$, and a \textit{Context Perceiver} (CP) denoted as $f_{\theta^{\text{CP}}}$. Each of these modules follows a similar architectural pattern with Perceiver Layers as the core component, as depicted in Figure \ref{fig:arch}. 

\vspace{-0.1in}
\paragraph{Vision Perceiver} The Vision Perceiver takes the image feature extracted with ViT, $\mathbf{V}^{(C_i)} \in \mathbb{R}^{T_i^{\text{IMG}} \times d}$, as input, and it undergoes a sequence of transformations. It starts with a self-attention layer to capture inherent visual relationships. Subsequently, cross-attention layers are employed to interact with the text token embeddings $\mathbf{U}^{(C_i)}$, allowing the model to enrich its understanding by integrating information from the text domain. Finally, the processed data passes through a feed-forward layer, resulting in an output denoted as $f_{\theta^{\text{VP}}}(\mathbf{V}^{(C_i)}) \in \mathbb{R}^{T_i^{\text{IMG}} \times d}, i=1,...,N$. Subsequently, we concatenate the outputs generated by the Vision Perceiver for each of the $N$ context samples, yielding a feature matrix of shape $\mathbb{R}^{(\sum^N_{i=1} T_{i}^\text{IMG}) \times d}$
\begin{align*}
    \mathbf{E}^{\text{IMG}} = \{f_{\theta^{\text{VP}}}(\mathbf{V}^{(C_1)}, \mathbf{U}^{(C_1)}), ..., f_{\theta^{\text{VP}}}(\mathbf{V}^{(C_N)}, \mathbf{U}^{(C_N)})\}, 
\end{align*}
where $\{\cdot\}$ denotes the concatenation operation. 

\vspace{-0.1in}
\paragraph{Language Perceiver} The Language Perceiver, on the other hand, takes as input the text token embeddings $\mathbf{U}^{(C_i)} \in \mathbb{R}^{T_i^{\text{TXT}} \times d}$, which encapsulate the linguistic content of the sample. The processing pipeline mirrors that of the Vision Perceiver, starting with a self-attention mechanism to capture textual relationships, followed by interactions with the visual embeddings $\mathbf{V}^{(C_i)}$ via a cross-attention layer to create a holistic understanding of the multimodal context. The final step involves a feed-forward layer,  resulting in an output denoted as $f_{\theta^{\text{LP}}}(\mathbf{U}^{C_i}) \in \mathbb{R}^{T_i^{\text{TXT}} \times d}$. Similarly, we concatenate the outputs of all Language Perceivers to obtain a feature matrix of shape $\mathbb{R}^{(\sum^N_{i=1} T_{i}^\text{TXT}) \times d}$:
\begin{align*}
    \mathbf{E}^{\text{TXT}} = \{f_{\theta^{\text{LP}}}(\mathbf{U}^{(C_1)}, \mathbf{V}^{(C_1)}), ..., f_{\theta^{\text{LP}}}(\mathbf{U}^{(C_N)}, \mathbf{V}^{(C_N)})\}. 
\end{align*}

\vspace{-0.1in}
\paragraph{Context Perceiver} The Context Perceiver takes the lengthy outputs from either the Vision Perceiver or the Language Perceiver as input and produces a condensed set of contextual representations. This is achieved by using a fixed number of learnable embeddings of size $\mathbb{R}^{\frac{M}{2} \times d}$ as the input of the Context Perceiver, where $M$ is typically significantly smaller than $L$. These embeddings undergo processing through a self-attention layer, followed by a cross-attention layer with the feature matrix from either the Vision Perceiver or the Language Perceiver, and a feedforward layer. Specifically, for the Vision Perceiver, the learnable embeddings are denoted as $\mathbf{H}^{\text{IMG}} \in \mathbb{R}^{\frac{M}{2} \times d}$, and the output after applying the Context Perceiver is denoted as:
\begin{align*}
    \tilde{ \mathbf{H}}^{\text{IMG}} = f_{\theta^{\text{CP}}}(\mathbf{H}^{\text{IMG}}, \mathbf{E}^{\text{IMG}}) \in \mathbb{R}^{\frac{M}{2} \times d}.
\end{align*}
Similarily, for the Language Perceiver,  the learnable embeddings are denoted as $\mathbf{H}^{\text{TXT}} \in \mathbb{R}^{\frac{M}{2} \times d}$, and the output after applying the Context Perceiver is denoted as:
\begin{align*}
    \tilde{ \mathbf{H}}^{\text{TXT}} = f_{\theta^{\text{CP}}}(\mathbf{H}^{\text{TXT}}, \mathbf{E}^{\text{TXT}}) \in \mathbb{R}^{\frac{M}{2} \times d}.
\end{align*}

Following this, $\tilde{\mathbf{H}}^{\text{IMG}}$ and $\tilde{\mathbf{H}}^{\text{TXT}}$ are concatenated, resulting in a contextual feature matrix $\tilde{\mathbf{H}} \in \mathbb{R}^{M \times d}$. This matrix is prepended to the beginning of the sequence, along with the text token embeddings of $q$ and the image embeddings of $q$ (transformed by the LLaVA multimodal projector). This integrated data serves as the input for the Large Language Model (LLM) to facilitate further processing.

\begin{table}[!t]
\centering
\small
\scalebox{0.87}{
\begin{tabular}{l|l|cc}
    \toprule
    Method & AVG. & IMG & TXT \\
    \midrule
    \midrule
    Human Average & 88.40 & 87.50 & 89.60 \\
    UnifiedQA~\cite{scienceqa} & 74.11 & 66.53 & 66.42 \\
    GPT-3 CoT~\cite{scienceqa} & 75.17 & 67.43 & 74.68 \\
    GPT-4 CoT~\cite{chameleon} & 83.99 & 71.49 & 82.65 \\
    LLaMA-Adapter~\cite{llamaadapter} & 85.19 & 80.32 & 83.72 \\
    MMCoT$_{\rm Base}$~\cite{mmcot} & 84.91 & 82.90 & 87.88 \\
    MMCoT$_{\rm Large}$~\cite{mmcot} & 91.68 & 88.80 & 95.26  \\

    LLaVA-7B~\cite{llava} & 89.28 & 87.32  & 90.96  \\
    LLaVA-13B~\cite{llava} & 90.90 & 88.00 & 89.49 \\
    \midrule
    \rowcolor{gray!20} \textbf{\caml-7B} & \textbf{91.32} & 89.24  & 93.21  \\
    \rowcolor{gray!20} \textbf{\caml-13B} & \textbf{92.03} & 89.94 & 93.84 \\
    \bottomrule
\end{tabular}}
\vspace{-0.1in}
\caption{Comparison with state-of-the-art methods on ScienceQA benchmark: \caml finetuned on \texttt{train} split and evaluated on \texttt{test} split. ``AVG.'' represents the average accuracy of all ScienceQA questions. ``IMG'' refers to the questions that include image contexts, while ``TXT'' refers to the questions without any images. }
\vspace{-0.15in}
\label{table:llava_scienceqa_test}
\end{table}

\subsection{Model Training} In the training process, we freeze the retriever and vision encoder, and train \caml by minimizing the following causal language modeling loss.
\begin{align*}
    \ell = -\sum_{i=1}^{|y|} \log p_{\theta} (y_i |\hat{y}_{1:i-1}, q, C_{1},...,C_{N}),
\end{align*}
where $\theta \leftarrow (\theta^{\text{LLM}}, \theta^{\text{VP}}, \theta^{\text{LP}}, \theta^{\text{CP}}, \mathbf{H})$ is the model trainable parameter ($\theta^{\text{LLM}}$ is the parameter of LLM);  $y_i$ is the ground-truth target, and $\hat{y}_{1:i-1}$ is the $i-1$ preceding tokens of output $y_i$;  Specifically, we use Vicuna-7B and Vicuna-13B as our backbone language model, resulting in two models \caml-7B and \caml-13B.

\section{Experiment}
Table \ref{table:llava_scienceqa_test},  Table \ref{table:instruction_tuning_12bench}, and Table \ref{table:compare_fs_models} provide a comprehensive summary of the performance achieved by \caml-7B and \caml-13B across various multimodal benchmarks. In comparison to previous state-of-the-art approaches, \caml exhibits noticeable improvement and establishes a new state-of-the-art. We elaborate on the experimental settings in the following sections and offer a comprehensive description (\eg, hyperparamter configuration) of the settings in the supplementary material.

\subsection{Multimodal Reasoning on ScienceQA}

\begin{table*}[!t]
\renewcommand\arraystretch{1.15}
\centering
\small
\scalebox{0.67}{
\begin{tabular}{lll|ccccc|clcccc}
    \toprule
    \textbf{Method} & LLM & Data & VQA$^{\rm v2}$ & GQA & VizWiz & SQA$^{\rm I}$ & VQA$^{\rm T}$ & POPE & MME & MMB & MMB$^{\rm CN}$ & SEED  & MM-Vet  \\
    \midrule
    \midrule
    IDEFICS-9B~\cite{obelics} & LLaMA-7B & 353M+1M& 50.9 & 38.4 & 35.5 & - & 25.9 & - & - & 48.2 & 25.2 & -  & - \\
    InstructBLIP~\cite{instructblip} & Vicuna-7B & 129M+1.2M & - & 49.2 & 34.5 & 60.5 & 50.1 & - & - & 36.0 & 23.7 & 53.4  & 26.2 \\
    Qwen-VL~\cite{qwenvl} & Qwen-7B & 1.4B+50M &78.8 & 59.3 & 35.2 & 67.1 & \textbf{63.8} & - & - & 38.2 & 7.4 & 56.3  & - \\
    Qwen-VL-Chat~\cite{qwenvl} & Qwen-7B & 1.4B+50M &78.2 & 57.5 & 38.9 & 68.2 & 61.5 & - & 1487.5 & 60.6 & 56.7 & 58.2  & - \\
    LLaVA-1.5~\cite{llava15} & Vicuna-7B & 558K+665K&78.5 & 62.0 & 50.0 & 66.8 & 58.2 & 85.9 & \textbf{1510.7} & 64.3 & 58.3 & 58.6  & 30.5 \\
    \rowcolor{gray!20} \textbf{\caml-7B}  & Vicuna-7B & 558K$^{\dagger}$+665K & \textbf{79.4} & \textbf{62.7} & \textbf{51.2} & \textbf{67.9} & 58.0 & \textbf{86.4} & 1506.9 & \textbf{66.9} & \textbf{60.6} & \textbf{60.4}  &  \textbf{32.2} \\
    \midrule
     IDEFICS-80B~\cite{obelics} & LLaMA-65B & 353M+1M & 60.0 & 45.2 & 36.0 & - & 30.9 & - & - & 54.5 & 38.1 & -  & - \\
    
    BLIP-2~\cite{blip2} & Vicuna-13B & 129M & 41.0 & 41.0 & 19.6 & 61.0 & 42.5 & 85.3 & 1293.8 & - & - & 46.4  & 22.4 \\
    InstructBLIP~\cite{instructblip} & Vicuna-13B & 129M+1.2M& - & 49.5 & 33.4 & 63.1 & 50.7 & 78.9 & 1212.8 & - & - & -  & 25.6 \\
    Shikra~\cite{shikra} & Vicuna-13B &600K+5.5M & 77.4 & - & - & - & - & - & - & 58.8 & - & -  & - \\
    LLaVA-1.5~\cite{llava15} & Vicuna-13B & 558K+665K & 80.0 & 63.3 & 53.6 & 71.6 & 61.3 & 85.9 & 1531.3 & 67.7 & \textbf{63.6} & 61.6  & 35.4 \\

    \rowcolor{gray!20} \textbf{\caml-13B}  & Vicuna-13B & 558K$^{\dagger}$+665K & \textbf{80.2} & \textbf{63.7} & \textbf{57.4} & \textbf{72.3} & 59.9 & \textbf{86.7} & \textbf{1588.7} & \textbf{70.2} & \textbf{63.6} & \textbf{62.3}  & \textbf{36.4} \\

    \bottomrule
\end{tabular}}
\vspace{-0.1in}
\caption{Comparison with state-of-the-art large multimodal models on 11 benchmarks (VQA$^{\rm v2}$~\cite{vqadataset}, GQA~\cite{gqa}, VizWiz~\cite{vizwiz}, SQA$^{\rm I}$~\cite{scienceqa}: ScienceQA-IMG, VQA$^{\rm T}$~\cite{textvqa}: TextVQA, POPE~\cite{pope}, MME~\cite{mme}, MMB~\cite{mmbench}: MMBench, MMB$^{\rm CN}$~\cite{mmbench}: MMbench-Chinese, SEED~\cite{seed}: SEED-Bench, MM-Vet~\cite{mmvet}). \caml achieves the best performance on 10/11 tasks. $^{\dagger}$ denotes BLIP558K-pretrained projector is initialized for instruction tuning.}
\vspace{-0.0in}
\label{table:instruction_tuning_12bench}
\end{table*}

ScienceQA, as introduced in~\cite{scienceqa}, serves as a multimodal benchmark designed for the task of science question answering. It comprises 21,000 multiple-choice questions, encompassing various domains including biology, physics, chemistry, Earth science, and more. We use the ScienceQA train split to build the datastore search index and for the model training. In specific, Vicuna-v1.3 is used as the LLM backbone, and CLIP-ViT-L-14~\cite{clip} is adopted as the visual feature encoder. By default, we retrieve three samples from the datastore as the contexts. We conduct evaluation of the ScienceQA on test split and adopt the same datastore (\ie, train split) search index for contextual samples retrieval.


We compare \caml with various baselines including UnifiedQA~\cite{unifiedqa}, GPT-4 CoT~\cite{chameleon}, LLaMA-Adapter~\cite{llamaadapter}, MMCoT$_{\rm Base}$~\cite{mmcot}, MMCoT$_{\rm Large}$~\cite{mmcot}, LLaVA~\cite{llava} 7B and 13B. 
As shown in Table~\ref{table:llava_scienceqa_test},  \caml surpass baselines by a noticeable margin. It is worth noting that our model attains the best performance on average, surpassing the previous state-of-the-art model MMCoT$_{\rm Large}$\footnote{\caml-13B attains the highest AVG scores on the ScienceQA leaderboard when GPT4-review is not used}. We also notice that, on the IMG questions, \caml-13B outperforms LLaVA-13B and MMCoT$_{\rm Large}$ by a wide margin, which underscores \caml's strong capability in handling challenging questions that incorporate scientific images. Furthermore, we conduct detailed ablative studies with the ScienceQA dataset using \caml. See Section~\ref{sec:ablation} for more details.

\begin{table}[!t]
\small
\centering
\scalebox{0.7}{
\addtolength{\tabcolsep}{-0.5em}
\begin{tabular}{l|cc|ccc}
    \toprule
    \multirow{3}{*}{\textbf{Method (shots)}} & \makecell{COCO \\ Caption} & Flickr30k & OKVQA & VQAv2 & Vizwiz  \\
    
     & CIDEr & CIDEr & Acc & Acc &  Acc  \\
    \midrule
    \midrule
    \textit{\textbf{Retrieval-augmented models:}} & & & & &  \\
    \ RA-CM3~\cite{racm3} (2) & 89.1 & - & - & - & -  \\
    \ ReViLM~\cite{revilm} (0) & 60.8 & 52.1 & - & -  & -   \\
    \ ReViLM~\cite{revilm} (2) & 77.2 & - & - & - &  -   \\
    \ ReViLM~\cite{revilm} (4) & 90.5 & - & - & - &  -   \\
    \ ReViLM~\cite{revilm} (8) & 90.2 & - & - & - &  -   \\
    \midrule
    \textit{\textbf{Zero/Few-shot models:}} & & & & &  \\
    \ Uni-Perceiver (0)~\cite{uniperceiver} & 109.8 & 41.2 & - & - & - \\
    \ Flamingo-9B (4) & 93.1 & 72.6 & 49.3 &  56.3 & 34.9   \\
    \ Flamingo-9B (32) & 106.3 & 72.8  &  51.0 &  60.4 & 44.0   \\
    \ Flamingo-80B (4) & 103.2  & 75.1 &  57.4 &  63.1  & 39.6   \\
    \ Flamingo-80B (32) & 113.8  & 75.4 & 57.8  &   67.6 & 49.8   \\
    \ KOSMOS-1 (4)~\cite{kosmos1}  & 101.7 & 75.3 & - &  51.8 & 35.3  \\
    \ MMICL (4)~\cite{mmicl}  & - & 72.0 & -  & 70.6 & 50.3  \\
    \midrule
    \rowcolor{gray!20} \textbf{\caml-7B} (3) & 111.4 & 82.7 & 64.7  & 79.4 & 51.2  \\
    \rowcolor{gray!20} \textbf{\caml-13B} (3) & \textbf{116.8} & \textbf{84.5} & \textbf{66.3}  & \textbf{80.2} & \textbf{57.4} \\
    \bottomrule
\end{tabular}}
\vspace{-0.1in}
\caption{Comparison with state-of-the-art zero/few-shot models including retrieval-augmented counterparts on Captioning and VQA tasks: \caml are the ones that trained in instruction tuning (Sec~\ref{sec:instruction_following}) and are evaluated with 3-shots. \caml-13B achieves the best performance on 5/5 tasks, even outperforming Flamingo-80B model \cite{flamingo} under 32 shots.}
\vspace{0.1in}
\label{table:compare_fs_models}
\end{table}

\begin{table}[!t]
\centering
\small
\scalebox{1.0}{
\begin{tabular}{l|lcc}
    \toprule
    Method & AVG. & IMG & TXT \\
    \midrule
    \midrule
    \caml-7B Baseline & \textbf{91.3} & \textbf{89.2} & \textbf{93.2} \\
    \ - \textit{w/o Perceiver} & 89.7 & 85.8 & 93.2 \\
    \ - \textit{w/o Vision Perceiver} & 89.8 & 87.4 & 92.1  \\
    \ - \textit{w/o Language Perceiver} & 90.0 & 87.7 & 92.0 \\
    \ - \textit{w/o Shared weights} &  91.3 & 89.1 & 93.2   \\
    \bottomrule
\end{tabular}}
\vspace{-0.1in}
\caption{Ablation Experiments on \caml perceiver components and shared-weights Context Perceiver: \caml-7B models are evaluated on ScienceQA-\texttt{test}.}
\vspace{-0.2in}
\label{table:ablation_contextmodel_design}
\end{table}

\subsection{Multimodal Instruction Tuning} \label{sec:instruction_following}
\caml can be enhanced by tuning to follow multimodal instructions to handle a more diverse set of multimodal tasks. To substantiate this, we perform instruction tuning following \cite{llava15} on LLaVA-665K, an instruction-tuning data mixture of LLaVA-1.5. Following \cite{llava15}, we use Vicuna-v1.5 as the LLM backbone and CLIP-ViT-L-14-336px as the visual encoder. The datastore is built using the exact training set, LLaVA-665K, without external injected information for a fair comparison. LLaVA-665K comprises 665K multimodal samples, and is made up of LLaVA-Instruct-158K, ShareGPT-40K~\cite{sharegpt}, VQAv2~\cite{vqadataset}, GQA~\cite{gqa}, OKVQA~\cite{okvqa}, OCRVQA~\cite{ocrvqa}, A-OKVQA~\cite{aokvqa}, TextCaps~\cite{textcap}, RefCOCO~\cite{refcoco} and VG~\cite{vg}. We adopt a training strategy in which each individual sample might have a distinct $N$, spanning a range from $1$ to $3$ and denote it as ``mixed-shots training". During inference, we retrieve three relevant supporting multimodal samples from the datastore index. To evaluate the model effectiveness, we traverse 11 comprehensive benchmarks, including VQAv2, GQA, TextVQA~\cite{textvqa}, MME~\cite{mme}, POPE~\cite{pope}, MM-Vet~\cite{mmvet}, MMBench~\cite{mmbench}, MMBench-CN~\cite{mmbench}, SEED-Bench~\cite{seed}, and Vizwiz~\cite{vizwiz}. Results on ScienceQA image (denoted as SQA$^{\rm I}$) is also reported following \cite{llava15}.

We compare \caml with large multimodal baselines such as  BLIP-2~\cite{blip2},  InstructBLIP~\cite{instructblip}, Shikra~\cite{shikra}, IDEFICS-9B~\cite{obelics}, IDEFICS-80B~\cite{obelics}, Qwen-VL~\cite{qwenvl},  Qwen-VL-Chat~\cite{qwenvl}, and LLaVA-1.5~\cite{llava15}. 
As shown in Table~\ref{table:instruction_tuning_12bench}, \caml-13B demonstrates the best performance across all 11 benchmarks, surpassing LLaVA-1.5, which uses the same training data. Notably, when employing Vicuna-13B as its backbone, \caml-13B outperforms BLIP-2, InstructBLIP-13B, Shikra, and LLaVA-1.5-13B. Additionally, \caml-7B delivers impressive results, outperforming all other baseline models except for LLaVA-1.5-13B on ten of the benchmark datasets. On POPE, \caml-7B achieves performance on par with that of \caml-13B. These encouraging observations showcase \caml's efficacy in harnessing multimodal contextual information and following multimodal instructions. In contrast to \citet{llava15}'s assertion that LLaVA-1.5 cannot handle multiple images due to the absence of instruction-tuning data and context length limitations, our design demonstrates feasibility of training a model to effectively process multiple images merely with the same training set and efficiently manage long context length.

\begin{figure*}[!t]
	\begin{center}
		\includegraphics[width=0.32\linewidth]{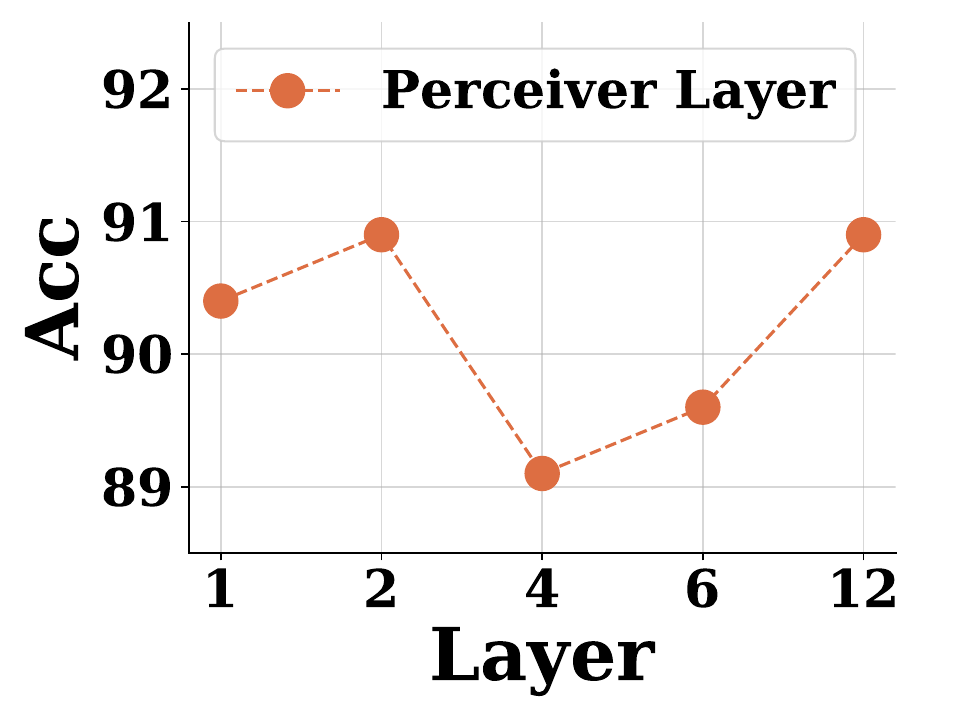}
        \includegraphics[width=0.32\linewidth]{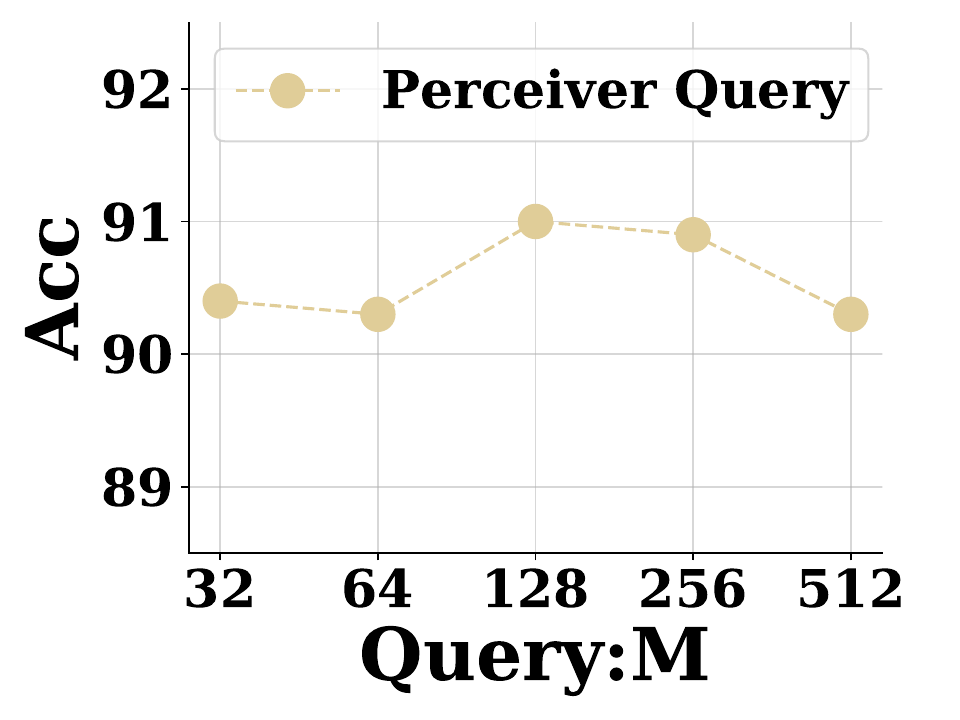}
        \includegraphics[width=0.32\linewidth]{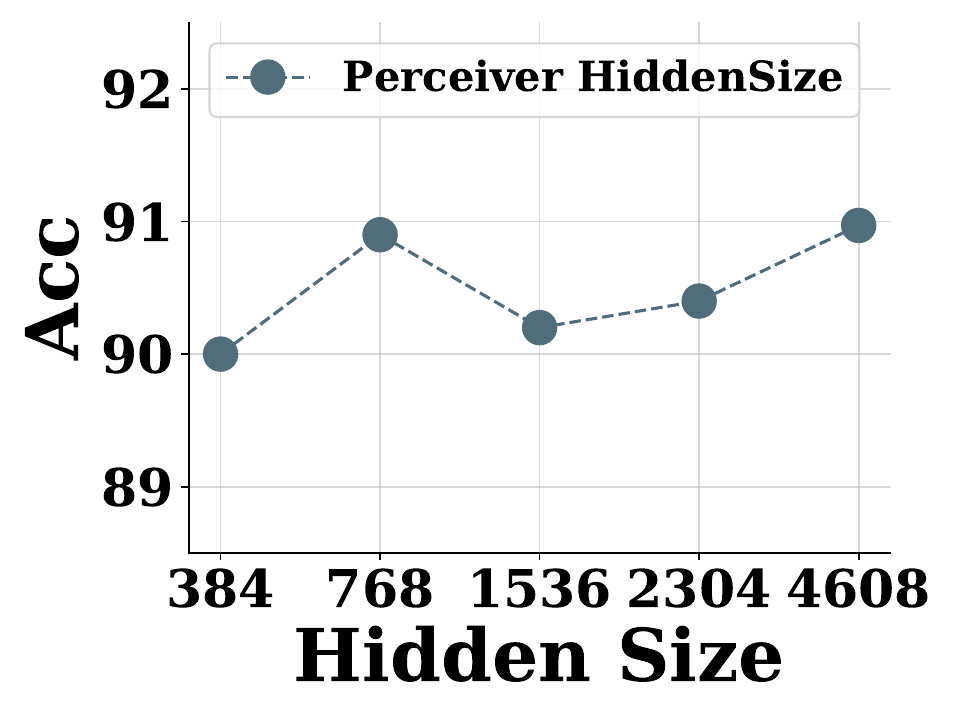}
	\end{center}
	\vspace{-0.1in}
	\caption{Ablation Experiments on \caml perceiver hyper-parameters: layers, query number $M$ and hidden sizes. \caml-7B with different settings are evaluated on ScienceQA \texttt{test}.}
	\label{fig:ablation_layer_query_hiddensize}
    \vspace{-0.1in}
\end{figure*}

\begin{figure*}[!t]
	\begin{center}
		\includegraphics[width=0.60\linewidth]{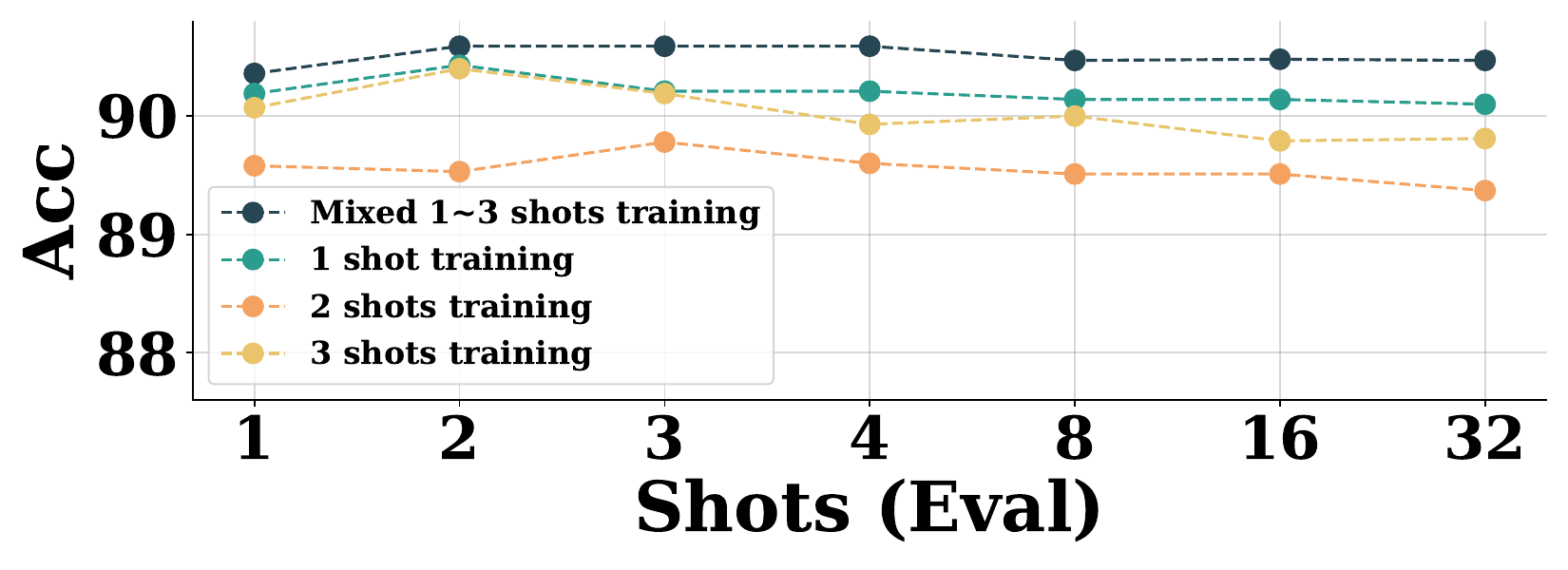}
        \hspace{0mm}
        \includegraphics[width=0.35\linewidth]{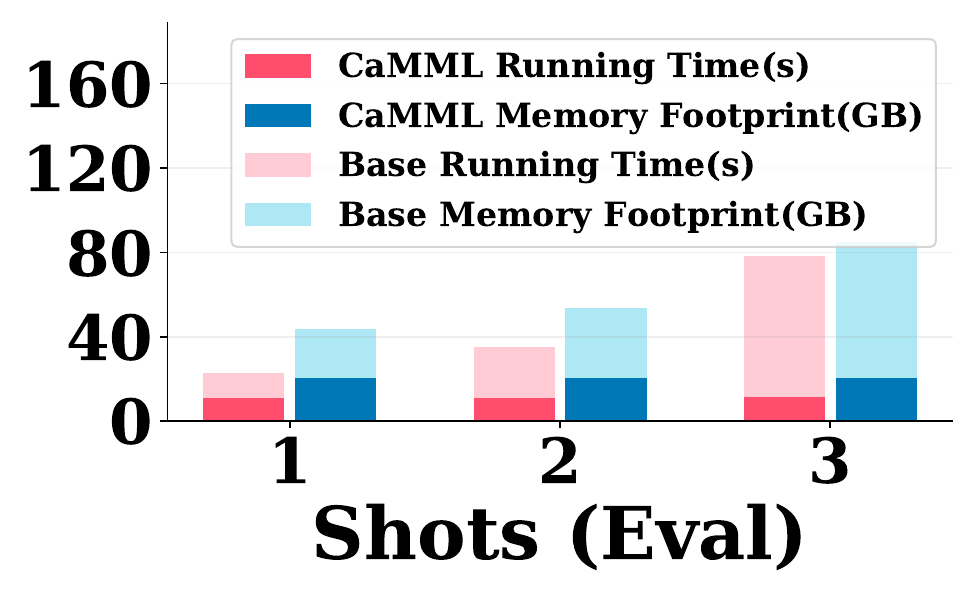}
	\end{center}
	\vspace{-0.1in}
	\caption{Ablation Experiments on \caml context number $N$. Left: different \caml models trained on $N$ shots are evaluated under 1$\sim$32 shots. Right: comparison between \caml and \caml without perceiver in terms of inference running time and memory footprint, the statistic is averaged on 100 samples from \caml-7B, which are tested on NVIDIA A100-80G GPU using ScienceQA dataset.}
    
    \label{fig:ablation_shot_inference}
	\vspace{-0.15in}
\end{figure*}

\subsection{Multimodal Few-shot Learning}
\caml essentially operates as a few-shot learning method. Few-shot models leverage the synergy of multiple intertwined images and texts to guarantee precise inference. Among these, retrieval-augmented models are widely recognized for their ability to retrieve highly similar samples, thereby improving few-shot prediction. In Table~\ref{table:compare_fs_models}, we conduct a comparison between \caml, retrieval-augmented models (RA-CM3~\cite{racm3} and ReViLM~\cite{revilm}), and other popular few-shot LLMs such as Flamingo~\cite{flamingo}, KOSMOS-1~\cite{kosmos1}, MMICL~\cite{mmicl} on tasks including visual captioning (COCO caption, Flickr30k), and visual question answering (OKVQA, VQAv2, Vizwiz). For \caml, we employ the same model described in Section \ref{sec:instruction_following}, with LLaVA-665K serving as the datastore. 

Several observations merit attention: (1) It is worth mentioning that while previous visual language models augmented with retrieval capabilities have been predominantly optimized for visual captioning tasks, \caml exhibits broad applicability across a wide range of tasks, including visual question answering and visual captioning. (2) Notably, \caml-13B outshines the significantly larger Flamingo-80B, highlighting \caml's superior efficiency and architectural effectiveness despite its smaller size. (3) Remarkably, \caml, even with a mere three-shot setup, surpasses models that necessitate 32 shots for equivalent tasks, underscoring its exceptional data efficiency and learning prowess.

\section{Model Analyses}

\subsection{Quantitative Analyses} \label{sec:ablation}
Here, we conducted an analysis of \caml, using the \caml-7B model for expeditious experiments, on the ScienceQA dataset. Our goal was to discern the significance of each module within the \caml architecture, as well as the influence of critical hyperparameters such as the number of layers, the selection of $M$, and the quantity of retrieved context samples. The default settings for \caml-7B baseline are as follows: the number of layers is set to 2 for all perceivers, $M=128$, $N=3$, and hidden size is set to 768. 
We also conduct experiments in Sec~\ref{sec:inference_cost} about \caml computation (inference speed \& memory footprint) compared with baseline approach, where all tokens are directly fed into LLMs.

\subsubsection{Contribution of Each Components}
We conducted ablation studies on the Perceiver, Vision Perceiver, Language Perceiver and Context Perceiver (shared weights or not) within \caml to evaluate the individual contributions of these components. The outcomes are documented in Table~\ref{table:ablation_contextmodel_design}. Our results highlight the critical role played by each component. The removal of Perceivers, in particular, leads to a marked deterioration in performance, indicating its significant influence on the model's overall effectiveness. Also, we have observed nearly identical results for both alternatives of the Context Perceiver, whether the weights are shared or not. Therefore, we prefer selecting the shared-weights option to save on computation.

\subsubsection{Impact of Hyperparameters}
In Figure ~\ref{fig:ablation_layer_query_hiddensize}, we report the performance by varying the number of layers, hidden sizes, and query number $M$. Our observations reveal several key insights: (1) Increasing the number of layers in \caml can have a detrimental effect on model performance. While we observe similar performance with both 2 layers and 12 layers, we opt for 2 layers due to its smaller model size; (2) Increasing the value of $M$ does not consistently lead to performance improvements; (3) Larger hidden sizes yield more favorable results in our analysis.

Furthermore, we investigated the impact of the number of context samples, $N$, used in the training stage as well as the inference stage. We conduct experiments where we varied the value of $N$ during training and inference, and the corresponding performance is reported in Figure~\ref{fig:ablation_shot_inference}. We observe that: (1) It is easy for \caml to accommodate a large number of shots; (2) Increasing the value of $N$ during inference does not consistently result in improved performance. Similar trends have been reported in previous works such as \cite{flamingo} and \cite{revilm}. One plausible explanation for this phenomenon could be that a longer context might introduce complexity and potentially convolute the inference process; (3) The ``mixed-shots" training strategy (Sec~\ref{sec:instruction_following}) has shown the potential to yield constant superior performance when compared to using a fixed value for $N$.

\subsubsection{Inference Cost}\label{sec:inference_cost}

As outlined in the model design, our context model adeptly manages large samples by condensing a significant number of raw tokens into a more streamlined representation. This efficient process leads to a faster forward pass in subsequent LLMs. We compared \caml with a baseline approach, where all context tokens are directly input into LLMs without \caml perceivers, and we present the inference speed and memory footprint in Figure~\ref{fig:ablation_shot_inference} (right). We have noticed that \caml only incurs negligible additional cost when increasing the number of context sample; As the number of context samples grows, the efficiency and memory benefits of \caml become more significant compared with the baseline method.

\begin{figure}[t]
    \centering
    \includegraphics[width=1.0\linewidth]{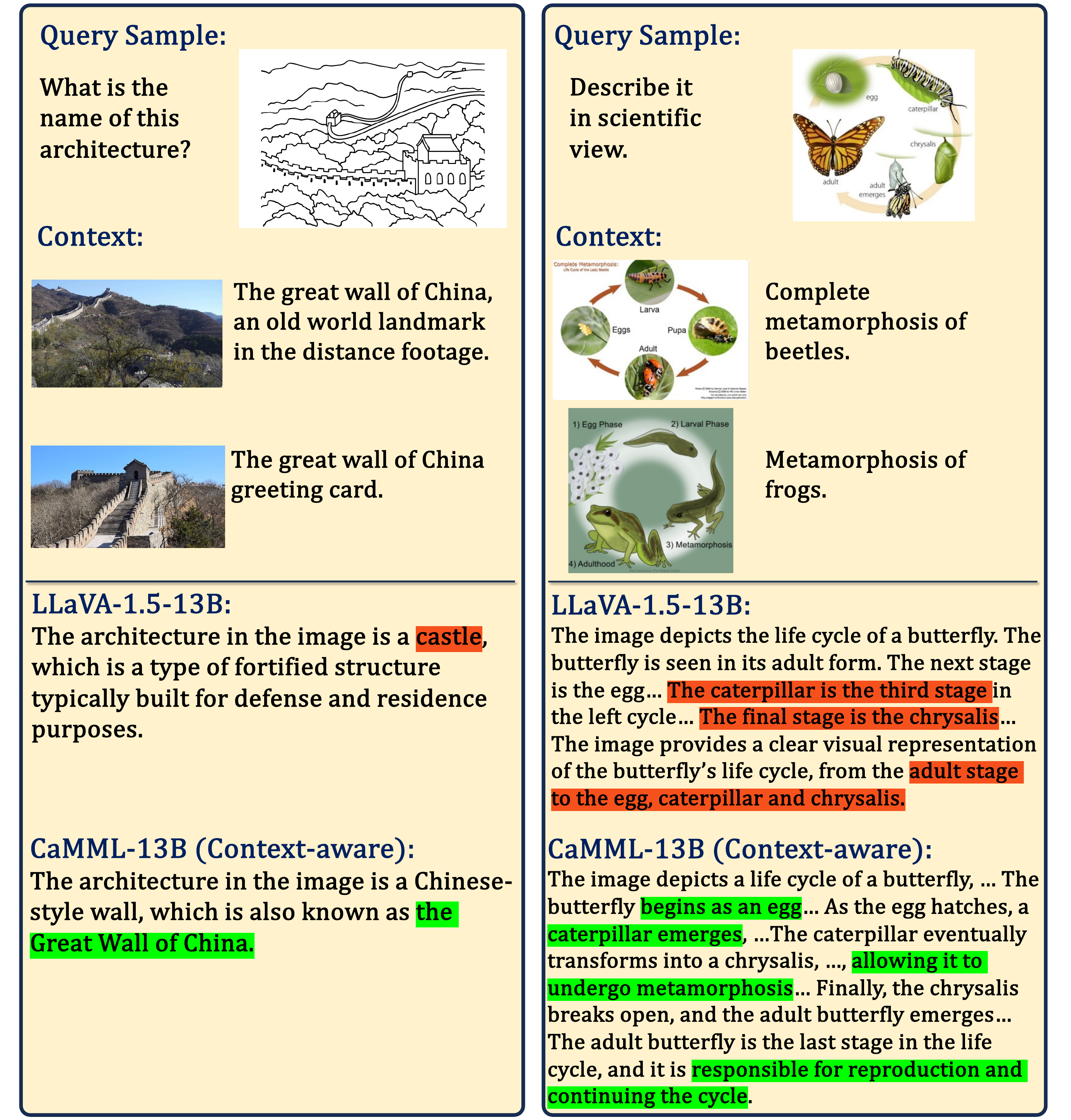}
    \vspace{-0.1in}
    \caption{Visualization of context-aware \caml \vs no-context LLaVA-1.5. Left: sketch drawing of the Great Wall. Right: depiction of metamorphosis of a butterfly.}
    \vspace{-0.1in}
    \label{fig:demo_context_aware}
\end{figure}

\begin{figure}[t]
    \centering
    \includegraphics[width=1.0\linewidth]{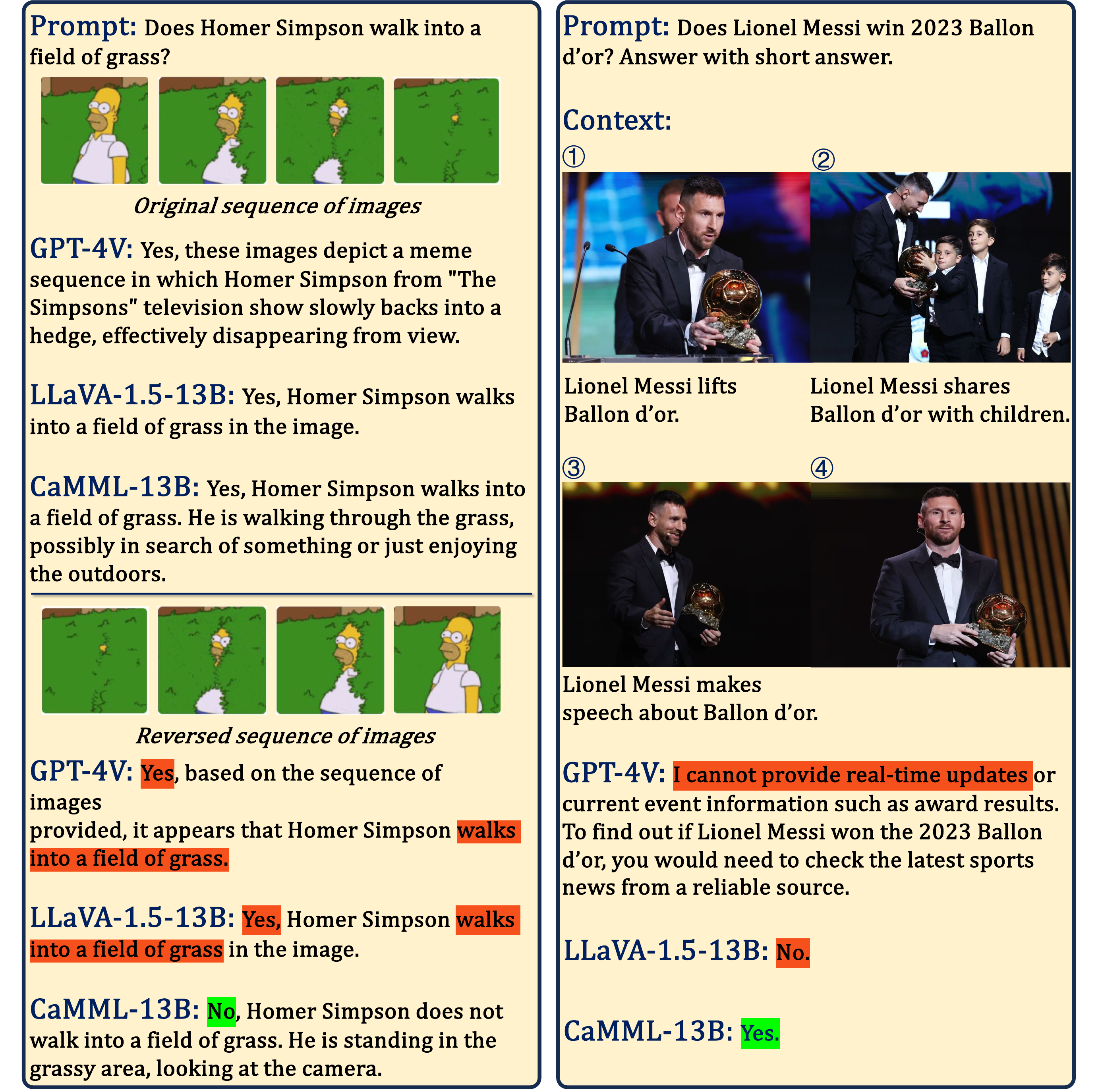}
    \vspace{-0.1in}
    \caption{Visualization of \caml \vs~GPT-4V $\&$ LLaVA-1.5. \caml demonstrates a strong understanding of contexual sequences. We directly input contexual sequences (Left: consecutive Homer Simpson video frames and Right: Lionel Messi winning Ballon d'or in October 2023.) to \caml w/o additional contexts.}
    \vspace{-0.in}
    \label{fig:demo_sequence}
\end{figure}

\subsection{Qualitative Analysis}

\subsubsection{The Importance of Context-Awareness}
\caml effectively processes a wide range of contextual inputs. In Figure \ref{fig:demo_context_aware}, we compare the response from \caml with context sample support and LLaVA-1.5, emphasizing the importance of context-awareness. The comparison illustrates that the inclusion of relevant context examples is crucial for ensuring the accuracy of answers. \caml can not only infer using highly relevant context samples (The Great Wall example), but also draw insights from analogous domain-specific samples (The Metamorphosis example) and conduct multimodal analogy-based learning~\cite{Antonietti2012}.

\subsubsection{Handling Image Sequences}
We also demonstrate that \caml possesses the ability to handle image sequences, even in the absence of explicit training for this specific task. In Figure \ref{fig:demo_sequence}, instead of retrieval, we consider the last image of the image sequence as the $q$ and the preceding images as context samples. A comparison with LLaVA-1.5 and GPT-4V in Figure \ref{fig:demo_sequence} on the QA task on image sequences reveals that while LLaVA-1.5 and GPT-4V struggle to fully comprehend the image sequences and learn from the priors, resulting in incorrect answers, \caml can effectively understand the image sequences and give accurate responses. More importantly, \caml can provide accurate inferences given access to the most up-to-date information regardless of LLM's internal knowledge, yet both LLaVA-1.5 and GPT-4V lack the capability to incorporate real-time updates of world knowledge.

\begin{figure}[!t]
    \centering
    \includegraphics[width=1.0\linewidth]{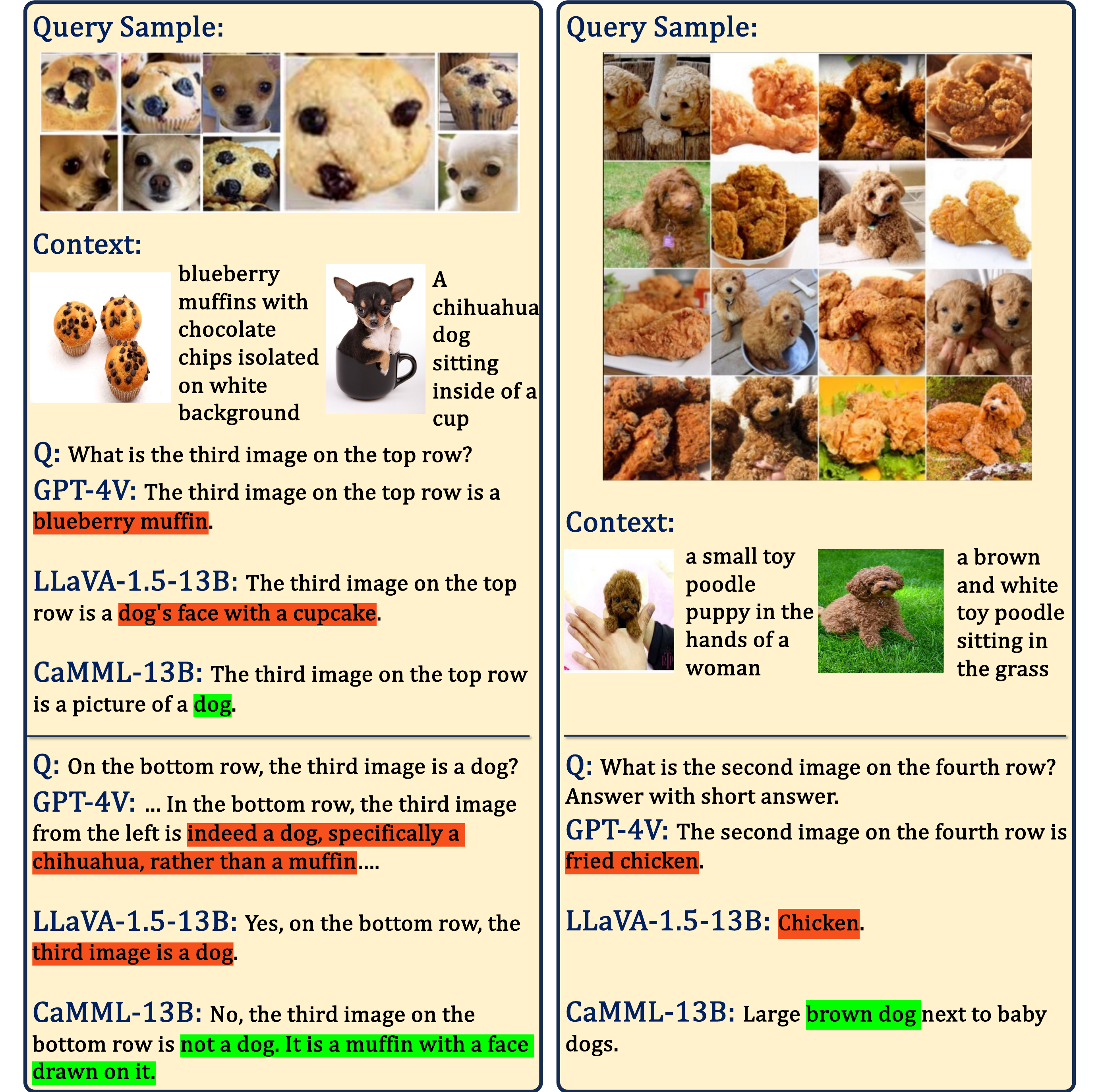}
    \vspace{-0.1in}
    \caption{Visualization of \caml \vs  GPT-4V $\&$ LLaVA-1.5. \caml effectively mitigates hallucinations and accurately identifies objects. On the left, there is an illusion of muffin and chihuahua. On the right, there is an illusion of fried chicken and labradoodle.}
    \vspace{-0.15in} \label{fig:demo_hallucination}
\end{figure}
\subsubsection{Tackling Multimodal Hallucination}
Previous research indicates that RAG is effective in reducing hallucinations in content generation~\cite{shuster2021retrieval}. Here, we aim to investigate if \caml can handle hallucinations in a multimodal setting. Figure \ref{fig:demo_hallucination} illustrates a multimodal QA task involving the recognition of similar objects: Muffins and Chihuahuas; Labradoodles and Fried Chicken. We treat the grid of images as a single image. It is evident that LLaVA-1.5 and GPT-4V face challenges in distinguishing between these objects, leading to hallucinated answers. In contrast, \caml accurately answers the questions with the groundness of relevant context samples.

\section{Conclusion}

We present a new methodology called \caml, a context-aware multimodal learner designed for the fine-tuning of large multimodal models.  With \caml, we build two multimodal models, \caml-7B and \caml-13B. \caml empowers them to draw insights from analogous,  domain-specific, and up-to-date context samples to make grounded inferences. Moreover, it employs a lightweight multimodal perceivers to seamlessly integrate these context samples, enabling an efficient processing of lengthy context tokens.  Our proposed \caml-13B model achieves state-of-the-art results across more than ten prominent multimodal benchmarks, surpassing previous methods by a substantial margin. These achievements underscore \caml's effectiveness in various multimodal applications.

\section{Limitations}

While \caml effectively integrates multimodal retrieval into large models, the presence of irrelevant examples may impede its performance.
In addition, \caml maintains a datastore for experimental analysis, yet it's unable to encompass all data domains, such as medical images. Consequently, when the data distribution in the datastore diverges significantly from user or test queries, \caml struggles to leverage its retrieval-augmented capability effectively.
Lastly, despite its proficiency in inference and memory utilization, \caml demands substantial training on large language models (similar to LLaVA) and relies on access to a sizable datastore to demonstrate its effectiveness.

\section{Potential Risks}
\caml entails certain potential risks, including:
Environmental unfriendliness and escalated computational costs, particularly noticeable with larger models.
Despite being entirely encapsulated within its pipeline, with users unable to access the datastore directly (limited to embedding and data entry index), there remains a possibility for users to utilize the \caml I/O API to instruct LLM to generate texts that may raise concerns or cause privacy leakage.

\bibliography{acl_camera_ready}

\appendix
\section{Extended Experiments}

\subsection{Experimental Setup}

\paragraph{Summary of Datasets and Benchmarks.}
ScienceQA~\cite{scienceqa} train split is used for ScienceQA finetuning and ablation study. In our instruction-tuning training, we adopt LLaVA-665K dataset, which contains LLaVA-158K, ShareGPT-40k~\cite{sharegpt}, VQAv2~\cite{vqadataset}, GQA~\cite{gqa}, OKVQA~\cite{okvqa}, OCRVQA~\cite{ocrvqa}, A-OKVQA~\cite{aokvqa}, TextCaps~\cite{textcap}, RefCOCO~\cite{refcoco} and VG~\cite{vg}. In our evaluation, VQAv2, GQA, TextVQA~\cite{textvqa}, MME~\cite{mme}, POPE~\cite{pope}, MM-Vet~\cite{mmvet}, ScienceQA, MMBench~\cite{mmbench}, MMBench-CN~\cite{mmbench}, SEED-Bench~\cite{seed} and Vizwiz~\cite{vizwiz} are considered as benchmarks. We also conduct evaluation on COCO Caption~\cite{cococap}, Flickr30k~\cite{flickr30k}, OKVQA~\cite{okvqa}, A-OKVQA~\cite{aokvqa}, and RefCOCO/+/g~\cite{refcoco}. In our qualitative visualization, we adopt another 2M datastore as the source of context examples, which comprised from external resources incorporates 2,348K multimodal samples, ranging from captioning with BLIP-LAION's 558K entries and Local Narrative~\cite{localnarrative}, to knowledge-based QA with KVQA~\cite{kvqa}, narrative-driven QA from VCR~\cite{vcr} and Visual7W~\cite{visual7w}, visual grounding via RefCOCOPlus~\cite{refcoco} and RefCOCOg~\cite{refcoco}, OCR from TextOCR~\cite{textocr}, along with the in-domain LLaVA-665K set.

\paragraph{Summary of Evaluation Metrics.}
We illustrate the evaluation metrics on Table~\ref{table:instruction_tuning_12bench}:
\begin{itemize}
    \item Accuracy: VQAv2, GQA, Vizwiz, SQA-Image, TextVQA, MMBench, MMBench-CN, SEED.
    \item GPT4-Assisted Evaluation Score: MM-Vet.
    \item F1 Score (POPE Paper Sec 5.1 Metrics): POPE.
    \item Figures in MME represent sum of the scores (which measures QA accuracy and accuracy+, illustrated in MME Paper~\cite{mme} Sec 2.2) of all MME perception subtasks, including existence, count, position, color, poster, celebrity, scene, landmark, artwork, and OCR. The full score of each subtask is 200, and that of all perception is 2000.
\end{itemize}

\paragraph{Summary of Pretrained Checkpoints.}
We utilize Vicuna-7B/13B~\cite{vicuna} as our foundation Large Language Model (LLM), ViT-L-14 architecture is used as the vision encoder. In detail, Vicuna-v1.3\footnote{\url{https://huggingface.co/lmsys/vicuna-7b-v1.3}, \url{https://huggingface.co/lmsys/vicuna-13b-v1.3}} and CLIP-ViT-L-14\footnote{\url{https://huggingface.co/openai/clip-vit-large-patch14}}~\cite{clip} initialization is for ScienceQA finetuning and Ablation studies, Vicuna-v1.5\footnote{\url{https://huggingface.co/lmsys/vicuna-7b-v1.5}, \url{https://huggingface.co/lmsys/vicuna-13b-v1.5}}, CLIP-ViT-L-14-336px\footnote{\url{https://huggingface.co/openai/clip-vit-large-patch14-336}}, and LLaVA-1.5 multimodal projector\footnote{\url{https://huggingface.co/liuhaotian/llava-v1.5-mlp2x-336px-pretrain-vicuna-7b-v1.5}, \url{https://huggingface.co/liuhaotian/llava-v1.5-mlp2x-336px-pretrain-vicuna-13b-v1.5}} is initialized for instruction finetuning. For the source of contexts, the ImageBind\footnote{\url{https://dl.fbaipublicfiles.com/imagebind/imagebind_huge.pth}}~\cite{imagebind}-Huge model is adopted to embed texts and images, computing their similarities for indexing top-k samples. 

\paragraph{Implementation Details.}\footnote{All the experiments are trained under Deepspeed Zero-3 FP16 configuration.}
In our ScienceQA finetuning, we train for 12 epochs, with batch size 4 per GPU and learning rate 2e-5. We illustrate the hyperparameter configurations here for ScienceQA finetuning in main paper Table 1: 
\begin{itemize}
    \item \textbf{\caml-7B:} \{number of query $M$=128, number of perceiver layer 2, perceiver layer hidden size 768, number of shots $N$=1\},
    \item \textbf{\caml-13B:} \{number of query $M$=256, number of perceiver layer 2, perceiver layer hidden size 4608, number of shots $N$=3\}.
\end{itemize}
In our instruction finetuning, we train for 1 epoch on 8GPUs, with batch size 8 per GPU and learning rate 2e-5, the LLM is activated as well as \caml perceivers, while the vision encoder and \caml retriever is frozen. We illustrate the hyperparameter configurations here for instruction finetuning in main paper Table 2\&3: 
\begin{itemize}
    \item \textbf{\caml-7/13B:} \{number of query $M$=128, number of perceiver layer 2, perceiver layer hidden size 768, mixed-training shots 1$\sim$3 and inferenced shots $N$=3\}.
\end{itemize}

\subsection{\caml Retrieval Setup}
We built \caml retriever in following steps:
\begin{itemize}
    \item Utilize ImageBind~\cite{imagebind} model to inference upon images and get corresponding visual embedding.
    \item Utilize Faiss~\cite{faiss} to build datastore index for visual embedding, and bind each data source (image and text) to the index, ensuring the right one-to-one retrieval.
    \item For each query, we forward with ImageBind model and compute similarity between query embedding and datastore index. We select top-k similarity samples as our contexts.
\end{itemize}
Note that, in our quantitative experiments, we compute query (text) embedding's similarity with (vision) datastore, while in qualitative analyses, we compute query (vision) embedding's similarity with (vision) datastore, to obtain relevant image information. 

\subsection{Experimental Results}

\begin{table}[!t]
\small
\centering
\scalebox{0.88}{
\addtolength{\tabcolsep}{-0.5em}
\begin{tabular}{l|cccc}
    \toprule
    \multirow{2}{*}{\textbf{Method (shots)}} & A-OKVQA & RefCOCO & RefCOCO+ & RefCOCOg   \\
    
     & Acc & Acc & Acc &  Acc  \\
    \midrule
    \midrule
    \textbf{\caml-7B} (3)  & 81.1 & 66.6 & 60.3 & 57.6   \\
    \textbf{\caml-13B} (3) & 82.0 & 70.6 & 65.9  & 60.5   \\
    \bottomrule
\end{tabular}}
\vspace{-0.1in}
\caption{\caml multimodal performance on A-OKVQA, Refcoco/+/g.}
\vspace{-0.1in}
\label{table:multimodal}
\end{table}

\subsubsection{Multimodal Task Performance} In addition to the results presented in the main paper tables, \caml demonstrates versatility in handling various multimodal tasks without requiring further fine-tuning. Table~\ref{table:multimodal} shows that \caml achieves exceptional performance on the Augmented Outside-Knowledge VQA (A-OKVQA) task with an accuracy of 82.0, and also exhibits good ability in localizing visual grounding tasks such as RefCOCO/+/g.

\subsubsection{Finegrained Evaluation} \caml is tested on MMBench and MMBench-CN to showcase each fine-grained ability such as Logic Reasoning (LR), Attribute Recognition (AR), RR (Relation Reasoning), Instance-Level Fine-Grained Perception (FP-S), Cross-Instance Fine-Grained Perception (FP-C), Coarse Perception (CP). According to Table ~\ref{table:finegrained_ability}, \caml-13B has achieved 8 out of 12 state-of-the-art results among large multimodal models.

\begin{table*}[!t]
\renewcommand\arraystretch{1.15}
\centering
\small
\scalebox{0.7}{
\begin{tabular}{ll|ccccccc|ccccccc}
    \toprule
    \multirow{2}{*}{\textbf{Method}} & \multirow{2}{*}{LLM} & \multicolumn{7}{c|}{MMBench-\texttt{dev}} & \multicolumn{7}{c}{MMBenchCN-\texttt{dev}} \\
    
     & & \textbf{Overall} & LR & AR & RR & FP-S & FP-C & CP & \textbf{Overall} & LR & AR & RR & FP-S & FP-C & CP \\
    \midrule

    MMICL~\cite{mmicl} & FLANT5-XXL & 67.9 & 49.2 & 71.6 & 73.0 & 66.7 & 57.2 & 77.2 & - & - & - & - & - & - & - \\ 

    mPLUG-Owl2~\cite{mplugowl2} & LLaMA2-7B & 66.5 & 32.2 & 72.4 & 60.9 & 68.6 & 60.1 & 79.4 & 59.5	&28.8&	64.8&	48.7&	60.1&	50.3&	76.0 \\ 
    
    SPHINX~\cite{sphinx} & LLaMA2-13B & 67.2 & 33.1 & 67.3 & 58.3 & 74.4 & 59.4 & 80.7 & 58.6	&21.2&	61.8&	43.5&	62.1&	58.7&	73.6\\ 

    LLaVA-1.5~\cite{llava15}-7B & Vicuna-7B & 63.0 & 26.3 & 68.8 & 53.0 & 67.2 & 56.6 & 76.4 & 57.4	&25.4&	58.8	&55.7&	55.3&	49.7&	75.7\\ 
        
    LLaVA-1.5~\cite{llava15}-13B & Vicuna-13B & 68.2 & 44.1 & 67.3 & 60.0 & 72.0 & 59.4	& 82.1 & 61.9&	36.4&	65.8&	49.6&	62.1&	59.4 &	75.0 \\ 
    
    \caml-7B & Vicuna-7B & 66.9 & 34.2 & 71.6 & 66.1 & 70.4 & 56.5 & 78.9 & 60.6 &27.1 &87.0 &55.7 & 58.0 & 57.3 & 77.0 \\ 
    \caml-13B & Vicuna-13B & \textbf{70.2} & 44.2 & \textbf{87.5} & 60.9 & 72.4 & \textbf{67.8} & \textbf{84.6} & \textbf{63.5} & \textbf{37.3} & \textbf{88.9} & 52.2 & \textbf{63.1} & \textbf{59.4} & \textbf{77.0} \\

    \bottomrule
\end{tabular}}
\caption{Experiments on fine-grained multimodal reasoning ability: \caml evaluated on MMBench and MMBenchCN, compared with other state-of-the-art methods. The categories include Logic Reasoning (LR), Attribute Recognition (AR), RR (Relation Reasoning), Instance-Level Fine-Grained Perception (FP-S), Cross-Instance Fine-Grained Perception (FP-C), Coarse Perception (CP).}
\label{table:finegrained_ability}
\end{table*}

\section{Additional Cases}

\begin{figure*}
    \includegraphics[width=1.0\linewidth]{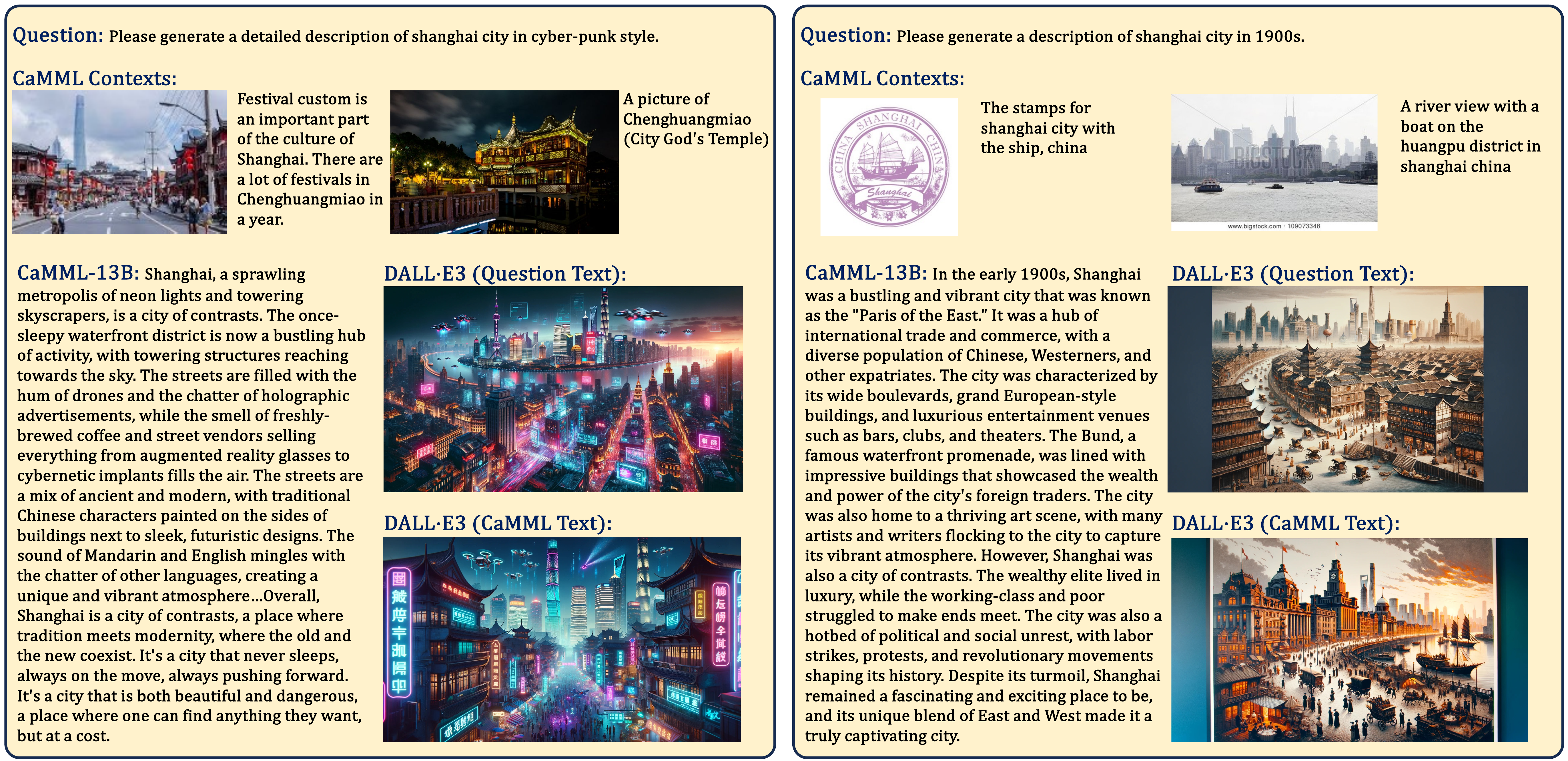}
    \caption{\caml capability of Image generation: Create a city view via DALL$\cdot$E3 API.}
    \label{fig:text_image_generation}
\end{figure*}

\begin{figure*}
    \includegraphics[width=1.0\linewidth]{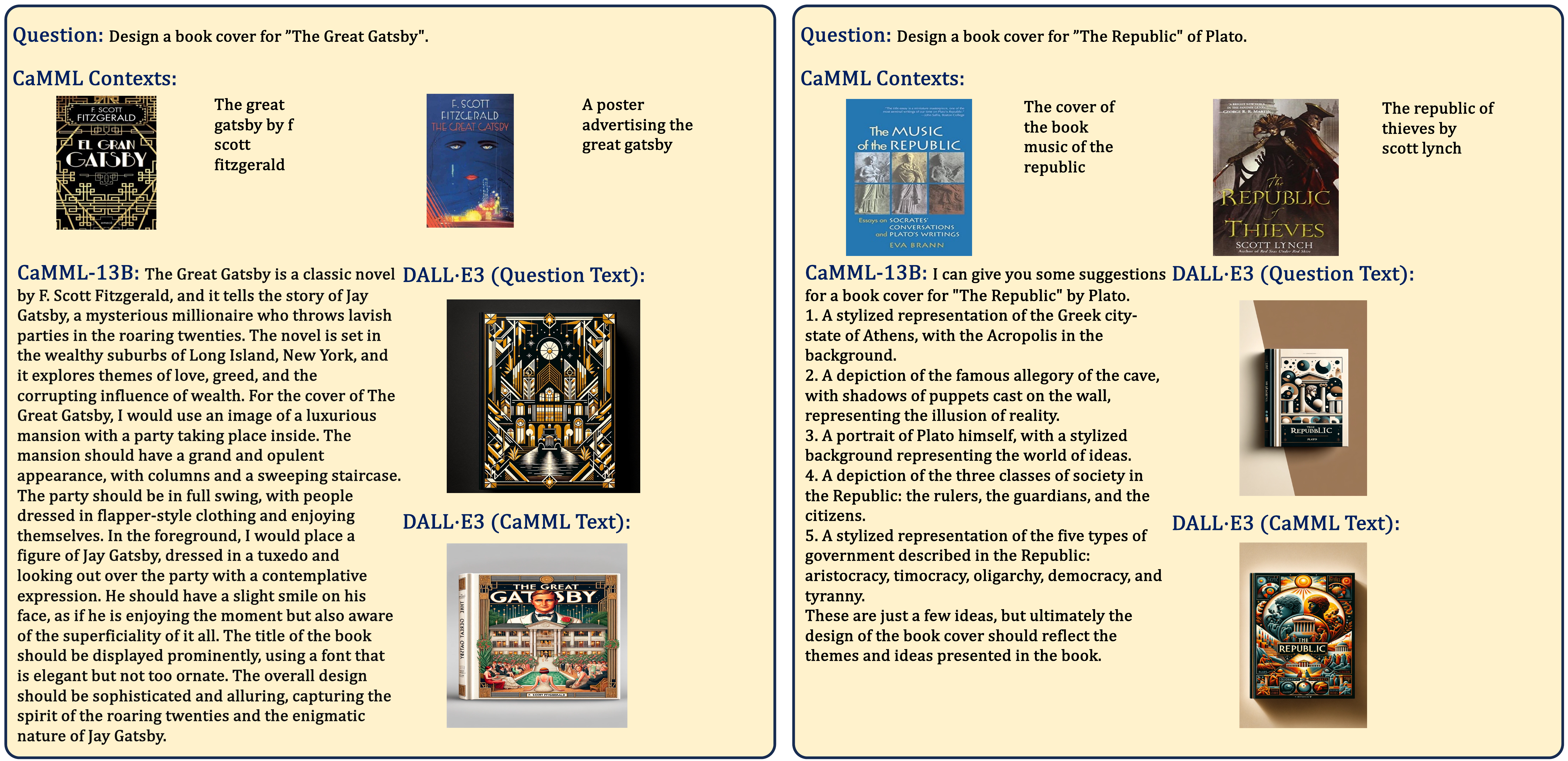}
    \caption{\caml capability of Image generation: Design a book cover via DALL$\cdot$E3 API.}
    \label{fig:text_image_generation2}
\end{figure*}

\begin{figure*}
    \includegraphics[width=1.0\linewidth]{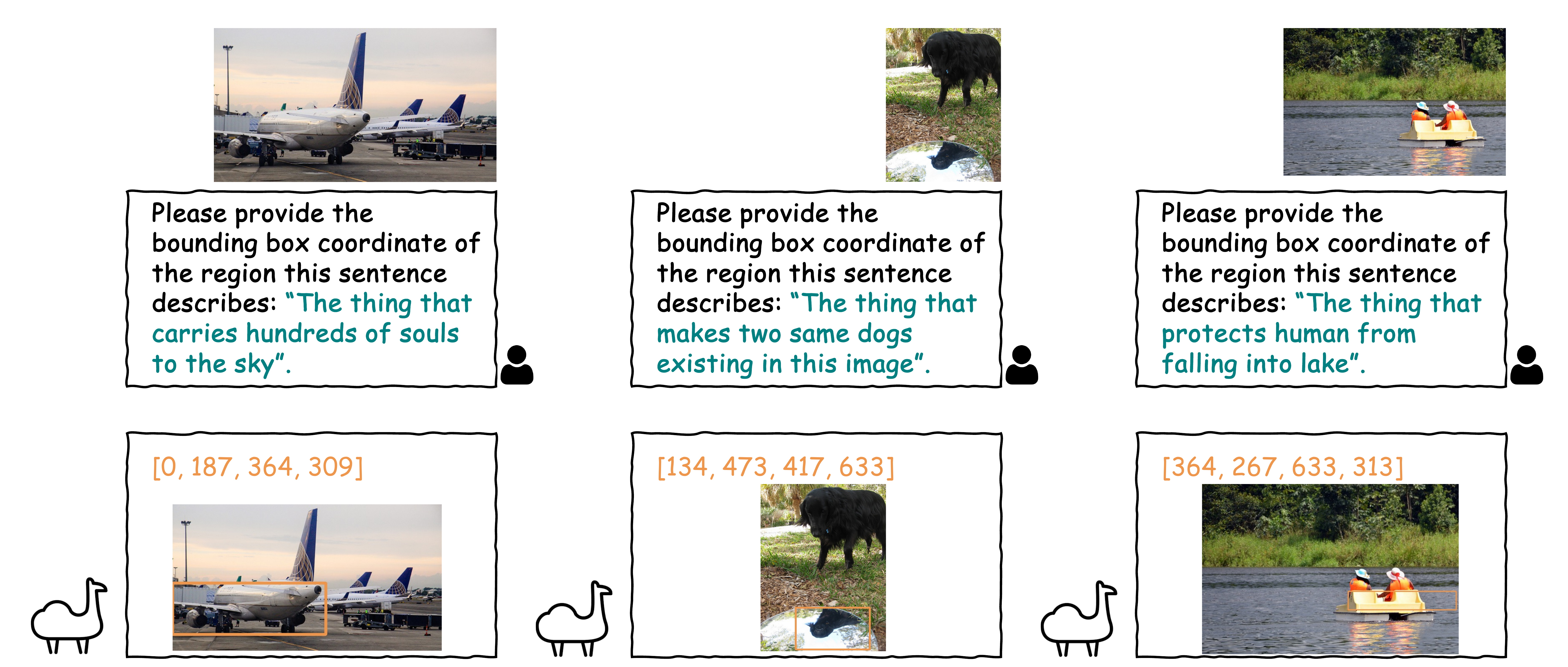}
    \caption{\caml capability of localization on implicit description.}
    \label{fig:demo_localization}
\end{figure*}

\paragraph{Image Generation}
\caml demonstrates its ability to generate reliable content based on a wealth of multimodal contextual sources. In Figure ~\ref{fig:text_image_generation}~\ref{fig:text_image_generation2}, we present an example of \caml's capability in prompting image generation. In the case, \caml is known for its ability to incorporate contextual elements into generating descriptions. For example, it can generate descriptions of traditional Chinese-style buildings in cyber-punk style. When using \caml, the generated images are likely to include a mix of old buildings with a modern style, based on the provided contextual samples. On the other hand, when using simple prompt without any contextual samples, the generated image is more likely to be limited to modern style skyscrapers only. 

\paragraph{Implicit-Description Localization}
\caml also highlights its strong localization capabilities, particularly in terms of reasoning the concept behind and high-level understanding. We utilize \caml to detect objects by providing implicit descriptions and allowing \caml to reason about the nature and location of these objects. As shown in Figure~\ref{fig:demo_localization}, \caml is able to identify the objects with implicit descriptions (such as airplane: ``The thing that carries hundreds of souls to the sky'' or mirror: ``The thing that makes two same dogs existing in this image'', \etc).

\end{document}